\documentclass[conference]{IEEEtran}
\IEEEoverridecommandlockouts

\usepackage{cite}
\usepackage{amsmath,amssymb,amsfonts}
\usepackage{algorithmic}
\usepackage{graphicx}
\usepackage{textcomp}
\usepackage{xcolor}
\def\BibTeX{{\rm B\kern-.05em{\sc i\kern-.025em b}\kern-.08em
    T\kern-.1667em\lower.7ex\hbox{E}\kern-.125emX}}

\def\namenospace{\textit{FedMoE-DA}}
\def\name{\textit{FedMoE-DA}\xspace}
\usepackage[ruled]{algorithm2e} 
\usepackage{comment}
\usepackage{booktabs}
\usepackage{multirow}
\usepackage{subfig}

\newcommand{\para}[1]{\noindent {\bf #1}} 

\begin{document}

\title{FedMoE-DA: Federated Mixture of Experts via Domain Aware Fine-grained Aggregation}

\author{
\thanks{Corresponding author: Xiaoxi Zhang (zhangxx89@mail.sysu.edu.cn)}
\IEEEauthorblockN{
Ziwei Zhan\IEEEauthorrefmark{1},
Wenkuan	Zhao\IEEEauthorrefmark{1},
Yuanqing Li\IEEEauthorrefmark{1},
Weijie Liu\IEEEauthorrefmark{2},
Xiaoxi Zhang\IEEEauthorrefmark{1}, \\
Chee Wei Tan\IEEEauthorrefmark{3},
Chuan Wu\IEEEauthorrefmark{2},
Deke Guo\IEEEauthorrefmark{4},
and Xu Chen\IEEEauthorrefmark{1}}
\IEEEauthorblockA{
\IEEEauthorrefmark{1}Sun Yat-Sen University, China\ \ \ \ \ 
\IEEEauthorrefmark{2}The University of Hong Kong, Hong Kong}
\IEEEauthorblockA{
\IEEEauthorrefmark{3}Nanyang Technological University, Singapore\ \ \ \ \ \IEEEauthorrefmark{4}National University of Defense Technology, China}}

\maketitle

\begin{abstract}
Federated learning (FL) is a collaborative machine learning approach that enables multiple clients to train models without sharing their private data. With the rise of deep learning, large-scale models have garnered significant attention due to their exceptional performance. However, a key challenge in FL is the limitation imposed by clients with constrained computational and communication resources, which hampers the deployment of these large models. The Mixture of Experts (MoE) architecture addresses this challenge with its sparse activation property, which reduces computational workload and communication demands during inference and updates. Additionally, MoE facilitates better personalization by allowing each expert to specialize in different subsets of the data distribution. To alleviate the communication burdens between the server and clients, we propose \name, a new FL model training framework that leverages the MoE architecture and incorporates a novel domain-aware, fine-grained aggregation strategy to enhance the robustness, personalizability, and communication efficiency simultaneously. Specifically, the correlation between both intra-client expert models and inter-client data heterogeneity is exploited. Moreover, we utilize peer-to-peer (P2P) communication between clients for selective expert model synchronization, thus significantly reducing the server-client transmissions. Experiments demonstrate that our \name achieves excellent performance while reducing the communication pressure on the server.
\end{abstract}


\section{introduction}
\label{sec:introduction}
Since the proposal of FL~\cite{fedavg}, it has garnered significant attention for its ability to collaboratively train machine learning models using distributed data without compromising user privacy. However, FL often encounters severe data heterogeneity, which affects model performance and can even prevent model convergence~\cite{fedavgconvergence}. With the advancement of deep learning technology, there is a trend towards utilizing large-scale models to enhance performance~\cite{largescalemodel}. Previous works~\cite{scalingNLM,scalingDL} indicate that a model's loss scales as a power-law with the number of model parameters, the volume of data, and the amount of computation used for training. Nevertheless, the computational resources at the client side are typically limited and highly heterogeneous~\cite{edgefl,adaptiveedgefl}, which contradicts the trend towards large-scale models.

Unlike dense models that activate all parameters for every input, sparsely activated models like Mixture of Experts (MoE) leverage conditional computation~\cite{sparsemoe}. MoE utilizes a gating network to partition the input space into several regions, with each expert model responsible for a specific region. This strategy lowers data complexity within each region, allowing expert models to employ simpler architectures while achieving good performance. 
Furthermore, by activating only portions of the model per sample, MoE enables larger model sizes without increasing computational load, thereby enhancing model performance without raising the computational budget. MoE is now widely used in natural language processing~\cite{switch} and computer vision~\cite{vmoe}.

FL suffers from two significant challenges, namely the system challenge and the statistical challenge, and the MoE architecture is a promising solution that helps mitigate both of these challenges for the following reasons.
First, as mentioned earlier, client-side computational and communication resources are typically limited. MoE can take advantage of sparse activation, allowing clients to train only a subset of the experts relevant to their local data rather than the entire model. This results in more efficient training and sparse communication, which mitigates system heterogeneity, and benefits devices with constrained resources.
Moreover, MoE can be scaled by adding more experts as needed without significantly increasing the computational burden on each client, enabling the model to handle larger and more diverse datasets efficiently.
In addition, the data distribution of clients is often heterogeneous. According to~\cite{fedmix}, the MoE architecture allows different experts to focus on various parts of the data or tasks. This enables the holistic model to capture the variances between datasets more effectively and generalize over a wider range of data distributions. Thus, deploying MoE-based models at distributed clients enhances the personalization performance of FL.

However, to the best of our knowledge, FL utilizing the {\em sparse} MoE is not well-studied, with only two notable works: FedMix~\cite{fedmix} and FedJETs~\cite{fedjets}. In these studies, each client maintains a model with an MoE architecture, comprising a gating network and several shared experts, and they leverage sparse activation to reduce communication overhead. 
We argue that these works have several shortcomings. First, when each communication round starts, each client needs to use the gating network and local data to prune less effective experts, introducing additional computation. Additionally, since the experts are shared across the system, clients can only choose from these shared models, which somewhat limits model personalization. Finally, although both algorithms suggest excluding some experts to reduce communication, multiple experts still need to be transmitted between the server and clients, increasing the communication burden on the server.

In order to mitigate the aforementioned shortcomings, we propose a novel FL system with an MoE architecture that can enhance {\em both the robustness and personalizability} of clients' models. First, drawing from previous works~\cite{representation1, representation2, representation3, pflmoe, fedjets}, which show that shallower layers in deep neural networks learn simple, low-level, and task-independent representations such as edges and textures, and considering that the MoE architecture struggles with high-dimensional data, we propose in our mechanism that all clients share an embedding model that is aggregated at the server to extract uniform and robust representations.
Secondly, to enhance the personalization capability of the client models, each client is equipped with its own unique gating network and experts. Moreover, to alleviate the communication burden of model aggregation, we utilize the relationship between gating network parameters and the selection patterns of experts to capture the correlation of experts among clients. We then leverage the P2P communication capabilities between clients to selectively transfer experts' parameters.
In addition, in order to further reduce the amount of transmission between the server and clients and to reduce the waiting time for aggregation, we periodically make aggregation decisions based on the latest round of information.

In summary, we make the following technical contributions:
\begin{itemize}
    \item We propose a new FL system utilizing the MoE architecture so as to enable task-independent embedding model aggregation for robustness enhancement and selective expert collaboration for personalizability improvement.  
    \item To reduce the communication overhead, we exploit the relationship between the gating network's parameters and expert selection patterns to capture expert correlations among clients without transmitting the entire model. We also propose an aggregation strategy that fully leverages clients' P2P communication capabilities and employ a periodic aggregation policy based on historical information. 
    \item We experimentally verify \namenospace's effectiveness, demonstrating its capability to achieve high model accuracy while minimizing server-client communication.
\end{itemize}
\section{related work}
\label{sec:related}

\subsection{Federated Learning with Non-iid Data}
\label{ssec:noniid}
Classical FL algorithms, such as FedAvg~\cite{fedavg}, aim to develop a global consensus model by aggregating updates from clients. However, the heterogeneous data across clients hinders the model convergence~\cite{fedavgconvergence}. Building on FedAvg, extensive efforts have been made to mitigate the adverse effects of Non-iid data, including penalized objective functions~\cite{fedprox,feddyn}, momentum-based variance reduction~\cite{scaffold,mime}, reweighted losses~\cite{fedir,ratio}, and contrastive losses~\cite{moon}.

Personalized FL allows clients to maintain their private models for better personalizability. Federated multi-task learning~\cite{mocha,fedem} trains client-specific models by capturing similarities among clients. Federated meta-learning~\cite{perfedavg, pfedme} aims to find an initial model that can be quickly adapted to local data. Knowledge distillation-based FL~\cite{distill, fedkd, feddistill} utilizes logits for information exchange, enabling heterogeneous models and efficient communication. Prototype-based FL~\cite{fedproto} performs collaborative model training by exchanging prototypes instead of gradients. Parameter decoupling~\cite{fedper, fedvf, fedpac, fedclassavg} views the model as a combination of a representation extractor and a classifier, developing separate aggregation strategies for each to achieve personalization while preventing overfitting.

However, most of these efforts rely on dense models, which limit the model size due to the limited computational resources available to clients, thereby affecting performance. In this paper, we explore the utilization of models with the MoE~\cite{moe, sparsemoe} architecture in FL, which are sparsely activated models that can increase the number of model parameters while keeping the computational load nearly unchanged.

\subsection{MoE-based Federated Learning}
\label{ssec:moe-fl}
In recent years, extensive research has explored the utilization of MoE in FL but has often overlooked the potential of leveraging sparse activation. PFL-MoE~\cite{pflmoe} and Zec et al.~\cite{zec} allow each client to train a personalized model and a global model, mixing outputs from both models via the MoE architecture to balance generalization and personalization performance. Based on the clustered FL algorithm IFCA~\cite{ifca}, Isaksson et al.~\cite{clustermoe} propose to mix the output of the local model and all the cluster models to enhance performance. 
Based on parameter decoupling, pFedMoE mixes the representations obtained by the local and global extractors, while FedRoD mixes the logits obtained by the local and global classifiers to balance the generalization performance as well as the personalization performance.
FedMix~\cite{fedmix} and FedJETs~\cite{fedjets} utilize models with the MoE architecture, enabling each client to exploit sparse activation property and data heterogeneity to selectively maintain an appropriate subset of experts for assembling, thereby reducing the computational load and communication volume.

Among these algorithms, the most similar to this paper are FedJETs and FedMix. However, both algorithms require clients to perform additional computation to select the appropriate subsets of experts and necessitate transmitting numerous parameters to the server, which increases the communication burden of the server. In contrast, our algorithm eliminates the need to evaluate experts' fitness and only requires transmitting the embedding model and the gating network, significantly reducing computation and communication overhead.

\section{Architecture}
\label{sec:model}

\subsection{Mixture of Experts}
\label{ssec:moe}

With the advances in deep learning technology, it is now widely recognized that larger models often perform better. The MoE architecture has gained significant attention for its ability to scale model size without increasing computational load. Typically, it consists of a gating function
$G(\cdot)$ as well as $K$ expert functions $\{E_i(\cdot)|i\in [K]\}$ with parameters $\{\mathbf{\Phi}^i | i\in [K]\}$ respectively. In the most typical setup, the gating network is an MLP network (whose parameters are denoted as $\mathbf{\Pi}\in \mathbb{R}^{n\times K}$) with softmax function, where $n$ is the input dimension of the gating network and experts. That is, given an input $\textbf{x}\in \mathbb{R}^{n}$:
\begin{align}
    G(\mathbf{x}) = \textit{SoftMax}(\mathbf{x} \cdot \mathbf{\Pi}), 
\end{align}
where the value of $G(\mathbf{x})\in \mathbb{R}^{K}$ indicates the selection scores for experts to process the input $\mathbf{x}$. We define $G_i(\mathbf{x})$ to be the value of the $i$\textsuperscript{th} dimension of the selection scores $G(\mathbf{x})$, which indicates the fitness of the $i$\textsuperscript{th} expert to the input $\mathbf{x}$. Subsequently, the input $\mathbf{x}$ would be sent to the experts for processing, and the outputs of these experts $\{ E_i(\mathbf{x}) | i \in [K]\}$ are combined to get the final output. That is,
\begin{align}
    \label{eq:MoE}
    \mathbf{y} = \sum_{i\in [K]} G_i(\mathbf{x})E_i(\mathbf{x}).
\end{align}

From this equation, we observe that when the selection score $G_i(\mathbf{x})$ is zero, the inference of the corresponding expert $E_i(\mathbf{x})$ can be omitted, which contributes to the sparsity of the MoE architecture. Additionally, we can further increase the sparsity of the activated model by selecting several experts with the highest scores, thereby reducing the computational burden.

Let's delve into the expert selection mechanism of the MoE architecture. As shown in Eq.~\ref{eq:MoE}, for each input $\mathbf{x}$, MoE uses $G(\mathbf{x})$ as the aggregation coefficient for the outputs of experts, and thus, $G(\mathbf{x})$ measures the relevance of each expert to the input $\mathbf{x}$. In other words, the function $G(\mathbf{\cdot})$ divides the entire representation space into several regions, and within each region, the most relevant experts are the same. We break the parameters of the gating network $\mathbf{\Pi} \in \mathbb{R}^{n\times K}$ into $K$ vectors $\{\mathbf{\pi}^i \in \mathbb{R}^{n} , i \in [K]\}$ by columns, and we refer to each vector as a proxy in the subsequent analysis, then the selection score of i\textsuperscript{th} expert $G_i(\mathbf{x})$
could be calculated as in Eq.\ref{eq:proxy}.
\begin{align}
    \mathbf{\Pi} := \begin{bmatrix}
        \mathbf{\pi}^1 & \mathbf{\pi}^2 & \cdots & \mathbf{\pi}^K
    \end{bmatrix};\\
    G_i(\mathbf{x})=\frac{\exp(\mathbf{x}\cdot \mathbf{\pi}^i)}{\sum_{j\in[K]}\exp(\mathbf{x}\cdot \mathbf{\pi}^j)}. \label{eq:proxy}
\end{align}

From Eq.~\ref{eq:proxy}, we could find that given input $\mathbf{x}$, 
$G_i(\mathbf{x})$ is positively correlated with the dot product of input $\mathbf{x}$ and the i\textsuperscript{th} proxy in the gating network $\mathbf{\pi}^i$. As a result, the probability of selecting the i\textsuperscript{th} expert $\mathbf{\Phi}^i$ increases as the angle between the input $\mathbf{x}$ and the proxy $\mathbf{\pi}^i$ decreases.

\subsection{Model Architecture: Applying Sparse MoE for FL}
\label{ssec:arch}
\begin{figure}[htp]
    \centering    \includegraphics[width=0.84\linewidth]{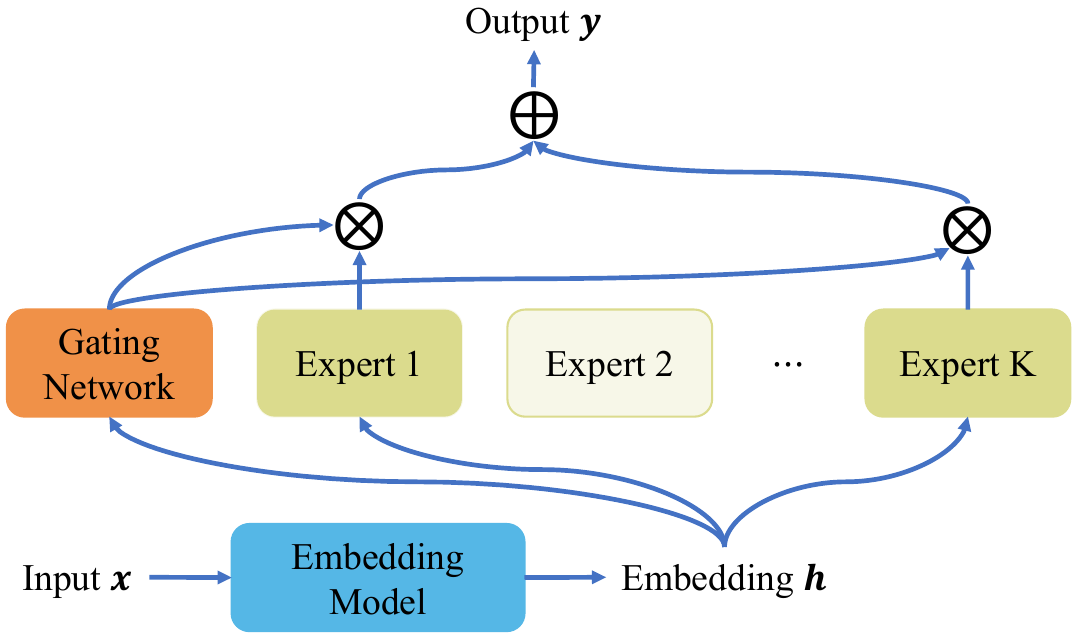}
    \caption{Model Architecture}
    \label{fig:arch}
\end{figure}

Previous studies \cite{representation1, representation2, representation3} have highlighted
that in deep neural networks, shallower layers typically learn general-purpose representations, while deeper layers capture representations that are relevant to the task and dataset. Additionally, the simplicity of the gating network has been found to limit its effectiveness with high-dimensional raw data \cite{pflmoe, fedjets}. To address these issues, we adopt the architecture depicted in Fig.~\ref{fig:arch}, comprising three main components: an embedding model, a gating network, and $K$ experts. During model inference, the embedding network first extracts a low-dimensional representation, which is then fed into the gating network to select the appropriate experts. Finally, the selected experts process the low-dimensional representation, and their outputs are aggregated using the gating network's output as weights.

\section{Algorithm Intuition and Design Details}
\label{sec:method}
To clarify our algorithm, we summarize the notation in Table~\ref{tab:notation} and demonstrate the pseudo-code in Algorithm~\ref{algo:fedmoe}. Note that for notational simplicity and clarity, we omit the subscripts indexing the communication rounds.

\begin{table}[!t]
\center
\normalsize
\begin{tabular}{cp{6.6 cm}}
    \toprule
    Notation & Description \\
    \midrule
    $N$ & Number of clients\\
    $T$ & Number of communication rounds\\
    $I$ & Aggregation matrix update interval\\
    $E$ & Number of updates in the local update phase\\
    $\eta$ & Learning rate for local updates\\
    $P$ & Number of experts requested by each expert\\
    $\mathbf{\Theta}_i$ & Local embedding model for the $i$\textsuperscript{th} client\\
    $\mathbf{\Theta}^t$ & Global embedding model at the $t$\textsuperscript{th} round\\
    $K_i$ & Number of experts in the $i$\textsuperscript{th} client\\
    $\mathbf{\Pi}_i$ & Gating model for the $i$\textsuperscript{th} client\\
    $\mathbf{\pi}_i^j$ & The $j$\textsuperscript{th} proxy of the $i$\textsuperscript{th} client\\
    $\mathbf{\Phi}_i^j$ & The $j$\textsuperscript{th} expert of the $i$\textsuperscript{th} client\\
    $\mathbf{W}_i$ & Local model for the $i$\textsuperscript{th} client, that is, $\mathbf{W}_i =\{ \mathbf{\Theta}_i, \mathbf{\Pi}_i, \{\mathbf{\Phi}_i^j \} \}$\\
    $A$ & Experts aggregation coefficient matrix\\
    \bottomrule
\end{tabular}
\caption{Summary of Notation}
\label{tab:notation}
\end{table}

\begin{algorithm}[t]
    \caption{FedMoE-DA}
    \label{algo:fedmoe}
    \LinesNumbered 
    \KwIn{$\mathbf{\Theta}^0$, $\{\mathbf{\Pi}_i\}$, and $\{\mathbf{\Phi}_i^j\}$, interval $I$, total number of local update epochs $E$, total number of communication rounds $T$}
    \KwOut{Personalized model $\{\mathbf{W}_i\} =\{ \mathbf{\Theta}^T, \{\mathbf{\Pi}_i\}, \{\mathbf{\Phi}_i^j \} \}$} 
    \For{communication rounds $t=1$~\textbf{\textup{to}}~$T$}{
        \ForEach{client $i\in [N]$ ~\textbf{\textup{in parallel}}}{
            Initialize local embedding model: $\mathbf{\Theta}_i \leftarrow \mathbf{\Theta}^{t-1}$;\\
            \For{local update epochs $e=1$~\textbf{\textup{to}}~$E$}{
            Perform local update: $\mathbf{W}_i = \mathbf{W}_i - \eta\nabla\mathbf{W}_i$;
            }
            Upload embedding model $\mathbf{\Theta}_i$ to the server;\\
            \If{$t \mod I == 1$}{
            Upload gating network $\mathbf{\Pi}_i$ to the server;}
            Request experts from other clients according to the aggregation matrix $A$;\\
            Aggregate experts using Eq.~\ref{eq:aggregate};\\
        }
        Server aggregates and broadcasts embedding model: $\mathbf{\Theta}^t = \frac{1}{N}\sum_{i\in [N]}\mathbf{\Theta}_i$;\\
        \If{$t \mod I == 1$}{
        Server computes and broadcasts the aggregation matrix $A$ according to Eq.~\ref{eq:weight};\\
        }
    }
\end{algorithm}

\subsection{Domain Aware Aggregation}
\label{ssec:aggregation}
Adopting the MoE architecture significantly increases the number of model parameters. Consequently, applying the vanilla FedAvg would lead to  a notable rise in communication between clients and the server, placing substantial pressure on the server and diminishing the algorithm's scalability. Therefore, a lightweight and efficient approach is essential to reduce the volume of server-client transmissions. To minimize the server-client transmissions and enhance the model's performance in terms of robustness and personalizability, we design different aggregation strategies for various  components.

\para{\bf{Robust embedding network.}} Given that shallower layers in models typically learn generic, task-independent representations, we expect uniform and robust representations from the embedding models across clients. Consequently, we adopt the vanilla FedAvg algorithm for the embedding model aggregation, i.e., we aggregate the embedding model on the server, and the aggregation rule is:
   \begin{align}
    \mathbf{\Theta}^t = \frac{1}{N}\sum_{i\in [N]}\mathbf{\Theta}_i.
\end{align}

\para{\bf{Shortcomings of previous MoE-based FL algorithms.}} Due to the variability of client data, different clients may have distinct preferences for experts. In FedMix~\cite{fedmix} and FedJETs~\cite{fedjets}, each client's model synchronizes with only a subset of the experts on the server, thus reducing extensive server-client communication. However, they have the following shortcomings. 
First, all clients share the experts, which limits model personalizability to some extent. Second, clients still need to synchronize multiple experts with the server, maintaining a heavy communication burden on the server. Third, they require client-side estimation of the relevance of experts, introducing additional computation.

\para{\bf{Indicators of expert relevance among clients.}} To address the aforementioned shortcomings, we allow clients to maintain their private gating networks and experts, and we propose an expert granularity domain-aware aggregation strategy that leverages the inspiration stated in Section~\ref{ssec:moe} as well as the P2P communication capabilities between clients. We assume that clients' gating networks are parameterized as $\{\mathbf{\Pi}_i\in \mathbb{R}^{n\times K_i} | i \in [N]\}$ and the experts are represented as $\{ \mathbf{\Phi}_i^j | i \in [N], j\in [K_i]\}$, where $\mathbf{\Pi}_i$ is the gating network of the $i$\textsuperscript{th} client, $\mathbf{\Phi}_i^j$ is the $j$\textsuperscript{th} expert of the $i$\textsuperscript{th} client, and $K_i$ is the number of experts in the $i$\textsuperscript{th} client.
To clarify the subsequent analysis and algorithm description, we reorganize them into the following form:
\begin{align}
    \hat{\mathbf{\Pi}} =& \begin{bmatrix}
        \mathbf{\pi}_1^1 & \cdots & \mathbf{\pi}_1^{K_1} & \cdots & \mathbf{\pi}_N^1 & \cdots & \mathbf{\pi}_N^{K_N}
    \end{bmatrix};\\
    \hat{\mathbf{\Phi}} =&\begin{bmatrix}
        \mathbf{\Phi}_1^1 & \cdots & \mathbf{\Phi}_1^{K_1} & \cdots & \mathbf{\Phi}_N^1 & \cdots & \mathbf{\Phi}_N^{K_N}
    \end{bmatrix};
\end{align}
where each column of $\hat{\mathbf{\Pi}}\in \mathbb{R}^{n \times (\sum_{i\in [N]}K_i)}$ is a proxy that corresponds to a specified expert located in the corresponding column of $\hat{\mathbf{\Phi}} \in \mathbb{R}^{|\mathbf{\Phi}| \times \sum_{i\in [N]}K_i}$. 

Based on the findings in section~\ref{ssec:moe}, the direction of proxies is pivotal in determining which inputs would activate corresponding experts. When two proxies exhibit high directional  similarity, 
it suggests that there may be a significant overlap in the regions for which each corresponding expert is responsible,
indicating that aggregating these two experts could be advantageous. Therefore, upon completing local updates, clients would send their gating networks to the server.
Then, we compute the similarity matrix $R=[r_{ij}]\in \mathbb{R}^{(\sum_{i\in [N]}K_i)\times(\sum_{i\in [N]}K_i)}$, where each element $r_{ij}$ represents the cosine similarity between the i\textsuperscript{th} and  j\textsuperscript{th} proxy in $\hat{\mathbf{\Pi}}$:
\begin{align}
    r_{ij} = \frac{\hat{\mathbf{\Pi}}[:,i] \cdot \hat{\mathbf{\Pi}}[:,j]}{\| \hat{\mathbf{\Pi}}[:,i]\| \|\hat{\mathbf{\Pi}}[:,j]\|}\label{eq:sim}.
\end{align}

We regard $r_{ij}$ as a substitute to the relevance of the i\textsuperscript{th} and j\textsuperscript{th} expert in $\hat{\mathbf{\Phi}}$. This allows us to capture the relevance among numerous experts without requiring clients to upload their experts to the server. Each client then requests the most relevant experts from other clients based on the similarity matrix, which is able to take advantage of the P2P communication capability among clients to reduce the communication burden of the server.

\para{\bf{Aggregation strategy for experts.}} To clearly illustrate the proposed aggregation strategy, without loss of generality, we focus on the i\textsuperscript{th} expert of the expert matrix $\hat{\mathbf{\Phi}}$. Each element of the $i$\textsuperscript{th} row of the similarity matrix (i.e., $r_{i\star}$) indicates the relevance of $i$\textsuperscript{th} expert with other experts. In order to minimize P2P communication between clients, we request only the most relevant $P$ experts for aggregation. Specifically, we define $\mathcal{S}_i$ is the set of experts considered by the $i$\textsuperscript{th} expert in the aggregation, that is:
\begin{align}
    \label{eq:requestset}
    \mathcal{S}_i = \{j |j\in [\sum_{k\in [N]}K_k], r_{ij} \ge top\_value(\mathbf{r}_{i\star}, P+1)\},
\end{align}
where the function $top\_value(\mathbf{r}_{i\star}, P+1)$ returns the $(P+1)$\textsuperscript{th} largest element of the input array $\mathbf{r}_{i\star}$.
Further, we utilize the similarity matrix to construct our expert aggregation rule: 
\begin{align}
    a_{ij} =&\begin{cases}
    \frac{\exp(r_{ij}/\tau)}{\sum_{k\in \mathcal{S}_i} \exp(r_{ik}/\tau)}& j\in \mathcal{S}_i,\\
    0 & otherwise,
    \end{cases}
    \label{eq:weight}\\
    \hat{\mathbf{\Phi}}[:,i] =& \sum_{j\in \mathcal{S}_i} a_{ij}\hat{\mathbf{\Phi}}[:,j],
    \label{eq:aggregate}
\end{align}
where the temperature parameter $\tau$ controls the smoothness of the distribution of  aggregation weights. A higher $\tau$ leads to a smoother distribution and a more robust aggregated expert, while a lower $\tau$ results in a sharper distribution, enhancing model personalization. In this paper, we set $\tau=1$. We define the aggregation matrix as $A=[a_{ij}]$, which contains all the information needed for expert aggregation.

\subsection{Historical Aggregation Matrix}
\label{ssec:historical}
Although the above approach reduces data transmission between the server and clients, it requires all clients to upload their gating network parameters. The server then calculates and broadcasts the aggregation matrix $A$. Clients can only perform P2P transmission after obtaining $A$, resulting in long waiting times.
We observed in our experiments (see Fig.~\ref{fig:interval}) that the staleness of the aggregation matrix does not significantly affect model performance. Therefore, it is unnecessary to compute a new aggregation matrix in each round. In order to eliminate waiting times and enhance training efficiency, we adopt a stale aggregation matrix for expert requests and aggregation. Specifically, each client maintains a stale aggregation matrix, allowing them to perform expert requests and aggregation based on this stale matrix during model training. This enables server-client communication to overlap with P2P communication among clients, eliminating waiting times. Every $I$ rounds, the server collects gating networks' parameters from  clients, computes, and broadcasts the updated aggregation matrix. This interval update of the aggregation matrix not only reduces server-client communication but also lessens the computational burden on the server.

\begin{figure*}[t]
    \centering \includegraphics[width=0.694\linewidth]{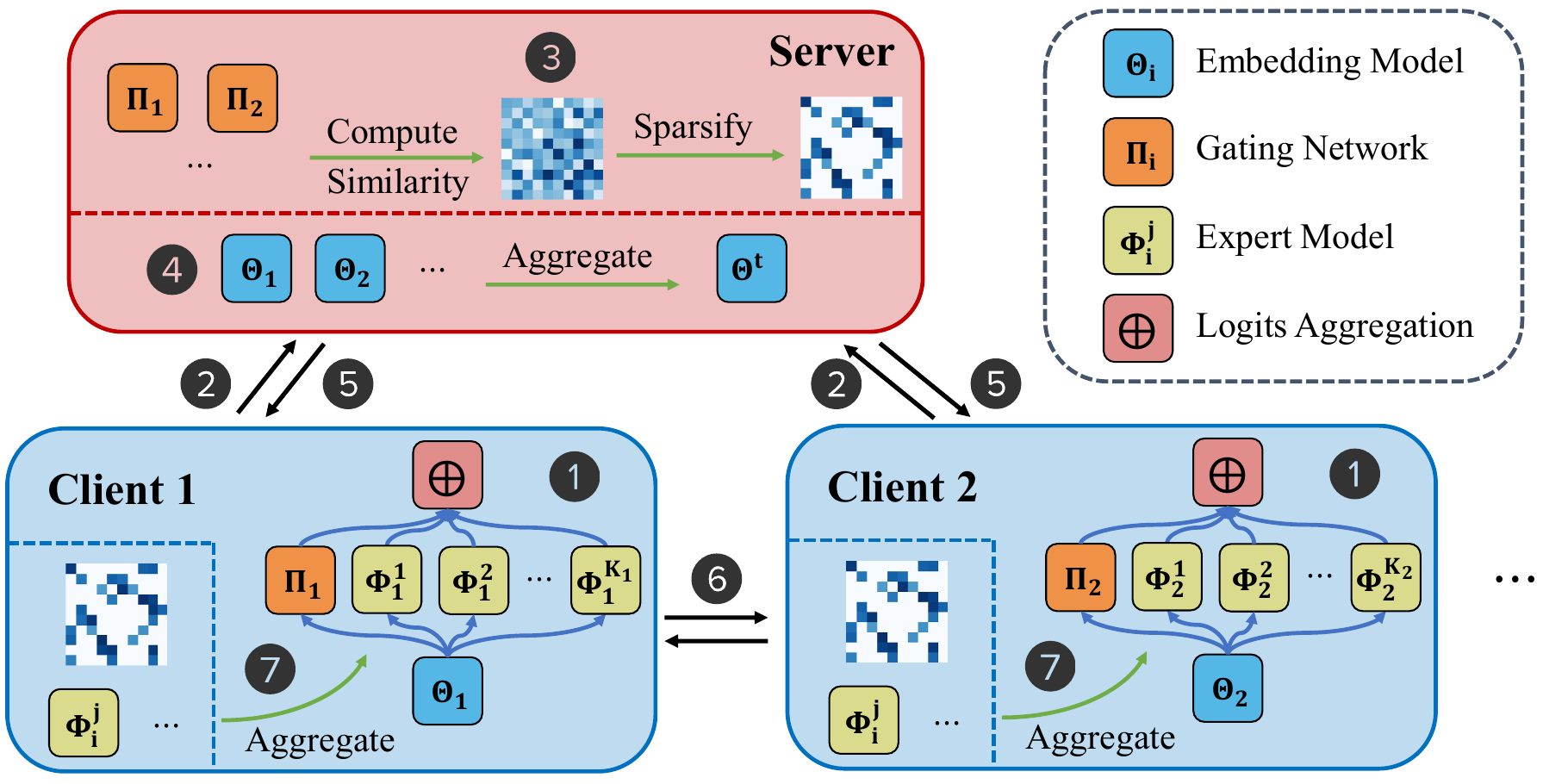}
    \caption{Workflow of \namenospace. (1)~Local update. (2)~Upload the local embedding model (and the gating network). (3)~Calculate the  aggregation matrix based on gating networks. (4)~Aggregate the  embedding model. (5)~Distribute the global embedding model (and the aggregation matrix). (6)~Transfer experts between clients according to the aggregation matrix. (7)~Aggregate experts using the requested experts and the aggregation matrix.}
    \label{fig:overflow}
\end{figure*}

\para{\bf{Algorithm workflow.}} We illustrate the workflow of \name in Fig.~\ref{fig:overflow}. During the $t$\textsuperscript{th} communication round, each client first performs $E$ epochs of local updates. Afterward, clients upload their local embedding models to the server. If it is a round for updating the aggregation matrix (i.e., $t \mod I == 1$), clients also upload their local gating networks to the server. The server then aggregates the local embedding models to obtain a new global embedding model and computes a new aggregation matrix based on the gating networks if needed. Then, the server broadcasts the global embedding model (and the aggregation matrix) to clients. Concurrently, clients request the most relevant $P$ experts from the corresponding clients for each expert and perform expert aggregation based on the stale aggregation matrix.

\para{\bf{Communication analysis.}} 
Using vanilla FedAvg, each client uploads and downloads the embedding model, gating network, and $K$ experts in each round, resulting in a total communication cost of $2(|\mathbf{\Theta}| + |\mathbf{\Pi}| + K |\mathbf{\Phi}|)$. 
In contrast, \name only requires the uploading and downloading of the embedding model in each round, along with the uploading of the gating network and the downloading of the sparse aggregation matrix every $I$ round.
Thus, the average server-client transmission for each communication round is $2 |\mathbf{\Theta}| + (|\mathbf{\Pi}| + \mathcal{O}(PK))/I$. Additionally, by utilizing the pre-trained model (see Sec.~\ref{ssec:pretrained}), we can avoid the aggregation of the embedding model, further reducing the server-client communication to $(|\mathbf{\Pi}| + \mathcal{O}(PK))/I$. However, \name necessitates expert P2P communication, with the average P2P transmission per client in each round being $PK|\mathbf{\Phi}|$. Overall,
\name effectively alleviates the communication pressure on the server through P2P communication, enhancing the robustness and scalability of the FL system.

\section{Experimental Results}
\label{sec:experiment}

\subsection{Experimental Set-up}
\label{ssec:setting}

\para{\textbf{Baselines:}} We compare our \name with the following algorithms: (1)~FedAvg~\cite{fedavg}, (2)~FedProx~\cite{fedprox}, (3)~Scaffold~\cite{scaffold}, (4)~FedProto~\cite{fedproto}, (5)~FedPer~\cite{fedper}, (6)~FedPAC~\cite{fedpac}, (7)~FedRoD~\cite{fedrod}, (8)~pFedMoE~\cite{pfedmoe}, (9)~FedMix~\cite{fedmix}, and (10)~FedJETs~\cite{fedjets}.

\para{\textbf{Models:}} We use a CNN model for our experiments, comprising two convolutional layers and two fully connected layers. For FedPer and pFedMoE, we designate the first two convolutional layers as the representation extractor and the last two fully connected layers as the classifier. For \name, we treat the first convolutional layer as the embedding model and the remaining layers as the experts.

\para{\textbf{Data partition:}} 
We evaluate \name using four classical computer vision datasets: MNIST, EMNIST, CIFAR-10, and SVHN. This paper focuses on label distribution skew, assuming each client has roughly the same amount of data, set to 500 samples per client. To simulate heterogeneity in label distributions, we use three commonly used data partitioning methods: homogeneous distribution, pathological distribution, and Dirichlet distribution. In the homogeneous distribution, label distributions are nearly identical across clients. In the pathological distribution, we adopt the  balanced and unbalanced two-class data partitioning methods from \cite{lotteryfl}, where each client has data from only two labels with the number of samples either being roughly the same (balanced) or inconsistent (unbalanced). The Dirichlet distribution~\cite{dirichlet} is controlled by a concentration parameter $\alpha > 0$, and the smaller the $\alpha$, the greater the degree of data heterogeneity among clients.

\para{\textbf{Hyperparameters:}} 
We assume a total of $N=50$ clients and employ stochastic mini-batch gradient descent with a learning rate of $\eta = 0.01$ and a batch size of $100$. The number of communication rounds is set to $T = 1000$, and the number of local training epochs per round is $E = 5$. For FL algorithms utilizing the MoE architecture (e.g., FedMix, FedJETs, and \name), we set the number of experts to $K_i=4(\forall i)$, activating only the expert with the highest selection score for each input to minimize the computational burden. For other unspecified hyperparameters, we use the values reported in the original papers.

\begin{table*}[htp]
\centering
\caption{Performance across different data distributions under CIFAR-10}
\label{tab:distribution}
\begin{tabular}{cccccccc}
\toprule
\multicolumn{2}{c}{\multirow{2}{*}{\#Method}}& \multirow{2}{*}{\textbf{Homogenous Distribution}} & \multicolumn{2}{c}{\textbf{Pathological Distribution}}  & \multicolumn{3}{c}{\textbf{Dirichlet Distribution}} \\ \cmidrule(l){4-8} 
\multicolumn{2}{l}{} & & balanced & unbalanced & $\alpha=0.1$ & $\alpha=1$ & $\alpha=10$ \\

\midrule
\multirow{6}{*}{Non-MoE}
&FedAvg & 57.76±0.52\% & 54.94±0.48\% & 54.97±1.20\% & 56.06±1.65\% & 58.70±1.57\% & 59.36±1.11\% \\
&FedProx & 57.74±0.53\% & 54.92±0.57\% & 54.91±1.14\% & 56.03±1.65\% & 58.67±1.58\% & 59.40±1.10\% \\
&Scaffold & \textbf{62.86±0.29\%} & 55.15±0.84\% & 54.67±1.14\% & 58.67±1.27\% & 62.68±1.48\% & \textbf{63.10±0.79\%} \\
&FedProto & 16.47±2.01\% & 54.16±1.87\% & 59.33±3.29\% & 27.49±2.17\% & 15.62±0.69\% & 17.08±1.51\% \\
&FedPer & 31.72±1.36\% & 81.69±1.10\% & 85.41±0.85\% & 79.02±3.77\% & 45.02±1.21\% & 34.01±0.72\%\\
&FedPAC & 45.27±3.58\% & 51.71±1.60\% & 75.00±0.00\% & 72.79±2.80\% & 37.02±0.97\% & 36.95±1.02\%\\
\midrule
\multirow{5}{*}{MoE}
&FedRoD & 45.51±1.08\% & \textbf{84.58±1.05\%} & \textbf{87.10±1.02\%} & 80.08±4.58\% & 54.29±0.66\% & 45.32±1.44\%  \\
&pFedMoE & 31.80±0.71\% & 78.40±1.01\% & 82.42±1.64\% & 76.01±2.62\% & 44.21±1.13\% & 34.85±0.63\% \\
&FedMix & 58.56±0.45\% & 25.25±0.38\% & 23.71±2.78\% & 28.98±5.30\% & 57.14±3.15\% & 59.39±0.69\%\\
&FedJETs & 51.15±1.59\% & 59.27±0.95\% & 59.00±2.19\% & 59.62±0.91\% & 52.80±3.04\% & 51.76±2.27\% \\
&\name & 60.66±1.31\% & 81.71±0.99\% & 83.79±1.80\% & \textbf{81.03±2.29\%} & \textbf{63.28±1.91\%} & 61.54±0.87\%\\ \bottomrule
\end{tabular}
\end{table*}

\begin{table*}[thp]
\center
\caption{Performance comparison under different datasets}
\label{tab:dataset}
\begin{tabular}{cccccc}
\toprule
\multicolumn{2}{c}{\textbf{\#Method}} & \textbf{MNIST} & \textbf{EMNIST} & \textbf{CIFAR-10} & \textbf{SVHN} \\ \midrule
\multirow{6}{*}{Non-MoE}
&FedAvg & 98.55±0.07\% & 89.94±0.02\% & 58.70±1.57\% & 83.78±0.38\% \\
&FedProx & 98.55±0.08\% & 89.96±0.08\% & 58.67±1.58\% & 83.77±0.40\%\\
&Scaffold & 98.48±0.11\% & 90.01±0.22\% & 62.68±1.48\% & 83.86±0.07\%\\
&FedProto & 13.07±8.22\% & 29.70±1.84\% & 15.62±0.69\% & 12.01±1.39\% \\
&FedPer & 92.70±0.38\% & 72.23±0.17\% & 45.02±1.21\% & 62.51±1.08\%\\
&FedPAC & 95.92±0.94\% & 85.20±1.53\% & 37.02±0.97\%  & 82.97±1.78\%  \\
\midrule
\multirow{5}{*}{MoE}
&FedRoD & 97.44±0.14\% & 84.67±0.26\% & 54.29±0.66\% & 79.30±0.81\%\\
&pFedMoE & 91.80±0.91\% & 74.90±0.38\%  & 44.21±1.13\% & 62.52±0.72\% \\
&FedMix & 98.52±0.07\% & 79.54±1.74\% & 57.14±3.15\% & 79.48±0.47\%\\
&FedJets & 98.40±0.08\% & 87.82±0.10\% & 52.80±3.04\% & 83.31±0.52\%  \\
&\name & \textbf{98.56±0.04\%} & \textbf{90.21±0.18\%} & \textbf{63.28±1.91\%} & \textbf{84.02±0.03\%} \\ \bottomrule
\end{tabular}
\end{table*}

\subsection{Performance Comparison on Various Data Partitions}
To validate \name under various data distributions, we conduct experiments on CIFAR-10 with the hyperparameters $P =5$ and $I=5$.
The results presented in Table~\ref{tab:distribution} show  that although \name does not achieve the highest accuracy in most distributions, it closely approaches the highest accuracy across all distributions. Scaffold excels in homogeneous distribution through synchronized updates, while FedRoD enhances personalization and robustness by integrating a personalized classifier for local optimization with a general classifier for broader knowledge. In contrast to FedRoD and Scaffold, which perform well only under extreme heterogeneity or isomorphism, \name demonstrates robustness across diverse data distributions. Table~\ref{tab:distribution} also demonstrates the advantages of using MoE to enhance FL performance. Additionally, to evaluate \namenospace's performance across different datasets, we conduct experiments on four datasets using a Dirichlet distribution with a parameter $\alpha=1$, as shown in Table~\ref{tab:dataset}. The results indicate that \name outperforms all baseline methods under this data distribution, highlighting the algorithm's strong performance across various datasets.

\subsection{Sensitivity Analysis}
\label{ssec:sensitivity}
To evaluate the impact of various hyperparameters on model training, we conduct experiments using the CIFAR-10 dataset, which is partitioned heterogeneously using a Dirichlet distribution with a parameter $\alpha =1$.

\begin{figure}[htbp]
    \centering
    \subfloat[Final Accuracy]{%
        \includegraphics[width=0.234\textwidth]{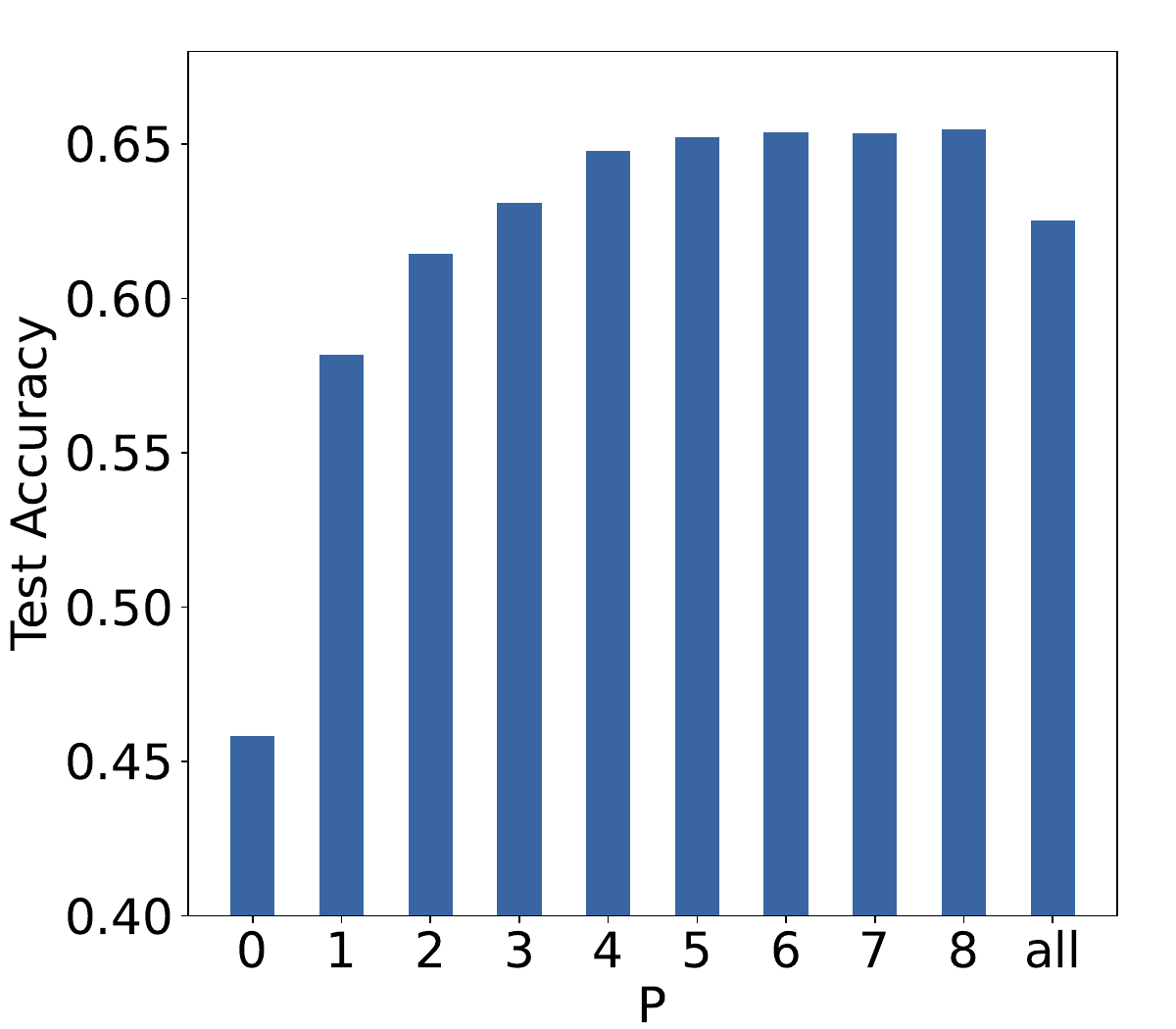}%
        \label{fig:k_acc}
    }
    \subfloat[Training Curve]{
        \includegraphics[width=0.234\textwidth]{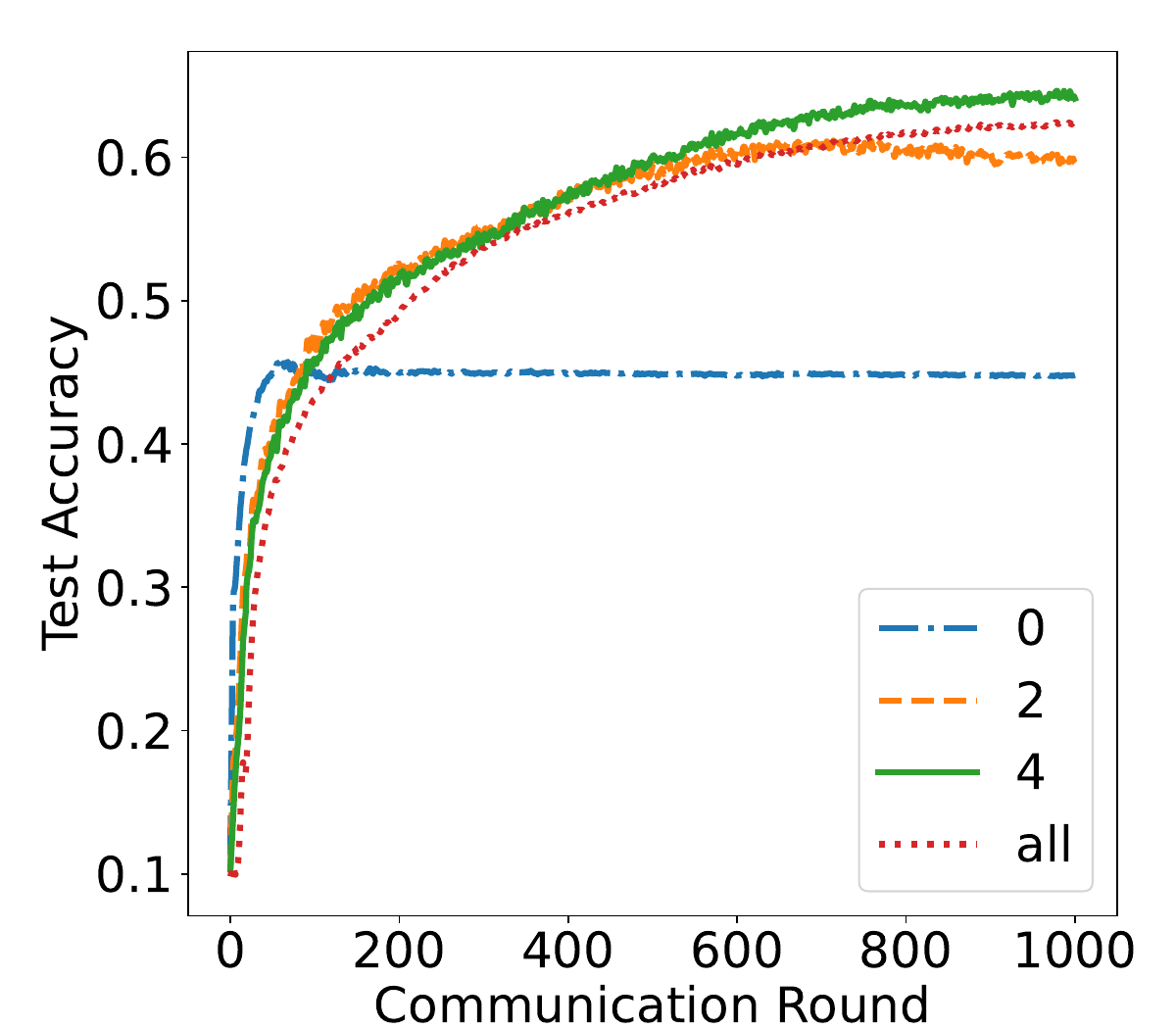}%
        \label{fig:k_curve}
    }
    \caption{The impact of the number of experts requested by each expert ($P$) on model training. A value of $0$ indicates that experts are not aggregated, while a value of ``all'' means that experts from all clients are considered during aggregation.}
    \label{fig:k}
\end{figure}

\para{\textbf{The number of experts requested by each expert $P$}}:
We conduct a sensitivity analysis of the hyperparameter $P$, with the interval between aggregation matrix updates set to $5$. Figure~\ref{fig:k_acc} illustrates the final prediction accuracy for various values of $P$. When $P$ is relatively small, accuracy improves as $P$ increases. This improvement is attributed to our meticulously designed aggregation matrix, which effectively captures and leverages correlations among experts to yield more robust and specialized models. Notably, when $P=0$, experts do not aggregate with others, resulting in significantly lower accuracy because the MoE architecture, with its substantial number of parameters, is prone to overfitting. Increasing the value of $P$ appropriately can effectively mitigate overfitting. However, aggregating all experts indiscriminately introduces excessive irrelevant knowledge, degrading model performance. Figure~\ref{fig:k_curve} illustrates the convergence rate for different values of $P$, revealing a slight deceleration in convergence as $P$ increases.

\begin{figure}[htbp]
    \centering
    \subfloat[Final Accuracy]{%
        \includegraphics[width=0.234\textwidth]{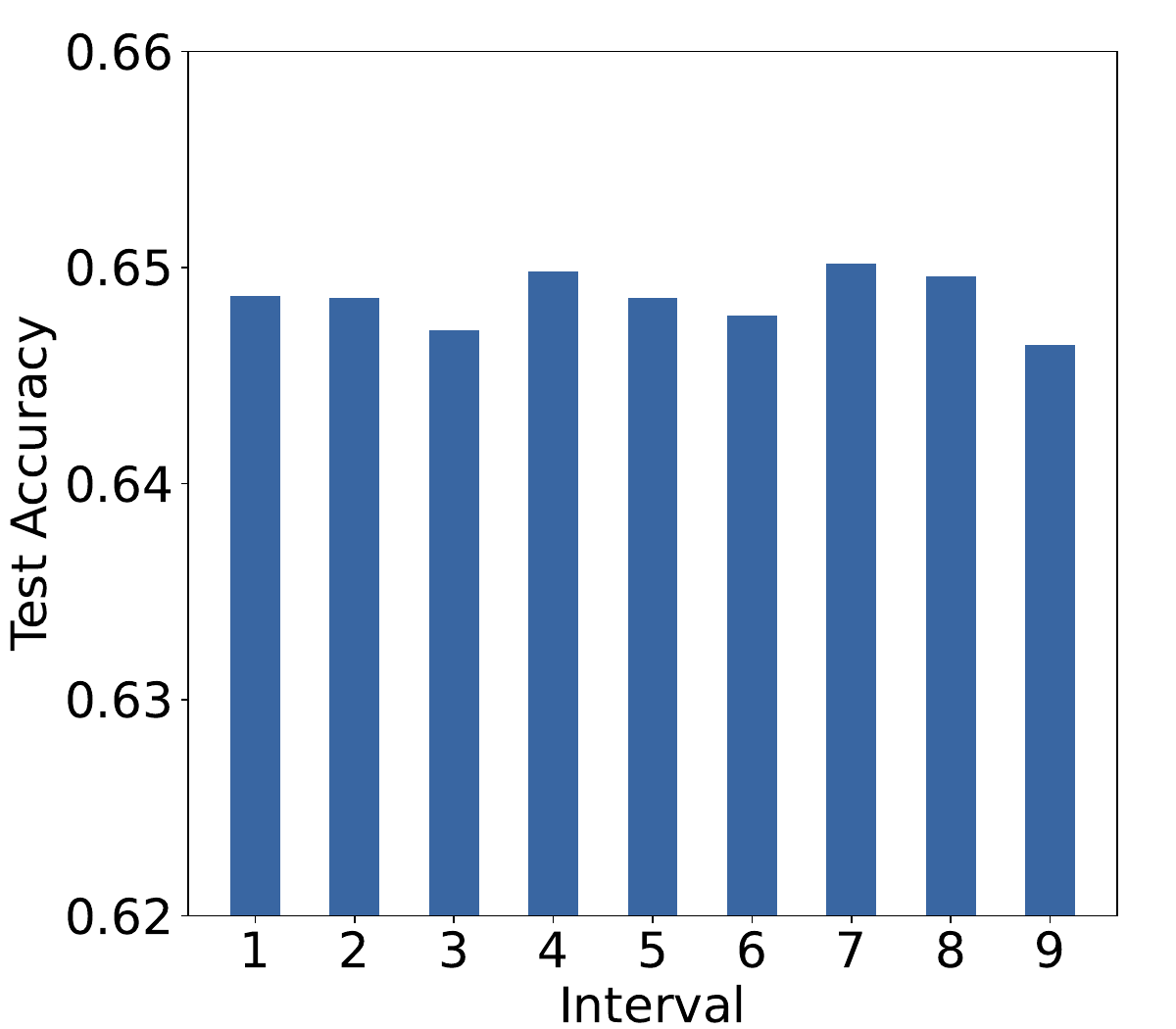}%
        \label{fig:gap_acc}%
    }
    \subfloat[Training Curve]{
        \includegraphics[width=0.234\textwidth]{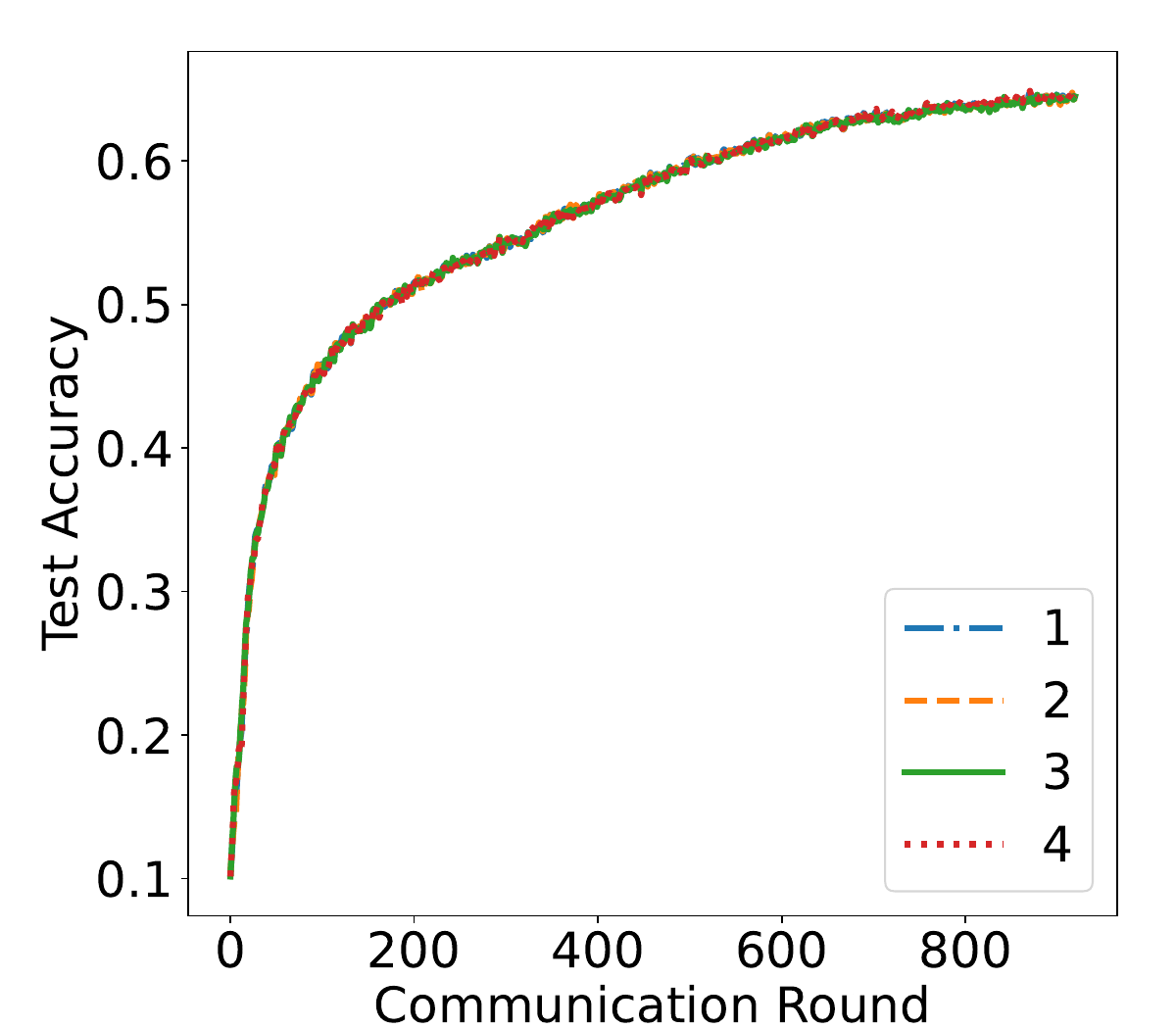}%
        \label{fig:gap_curve}
    }
    \caption{The effect of the interval rounds between the updates of the aggregation matrix on the model training.}
    \label{fig:interval}
\end{figure}

\para{\textbf{Interval between updates of the aggregation matrix $I$}}: We conduct a sensitivity analysis of the update interval $I$, with the number of experts requested by each expert set to $5$. Fig.~\ref{fig:gap_acc} illustrates the final prediction accuracy while Fig.~\ref{fig:gap_curve} presents the convergence rate across various values of $I$. The results indicate that the performance of the model is not sensitive to $I$. 
This stability allows for a larger interval between aggregation matrix updates, significantly reducing server-client communication and computational overhead. Specifically, fewer transfers of gating networks and the aggregation matrix lower communication costs, while a reduced update frequency of the aggregation matrix minimizes the computational burden, leading to a more efficient FL system.

\subsection{Utilizing Pre-trained Embedding Model}
\label{ssec:pretrained}
\begin{figure}[htbp]
    \centering
    \includegraphics[width=0.234\textwidth]{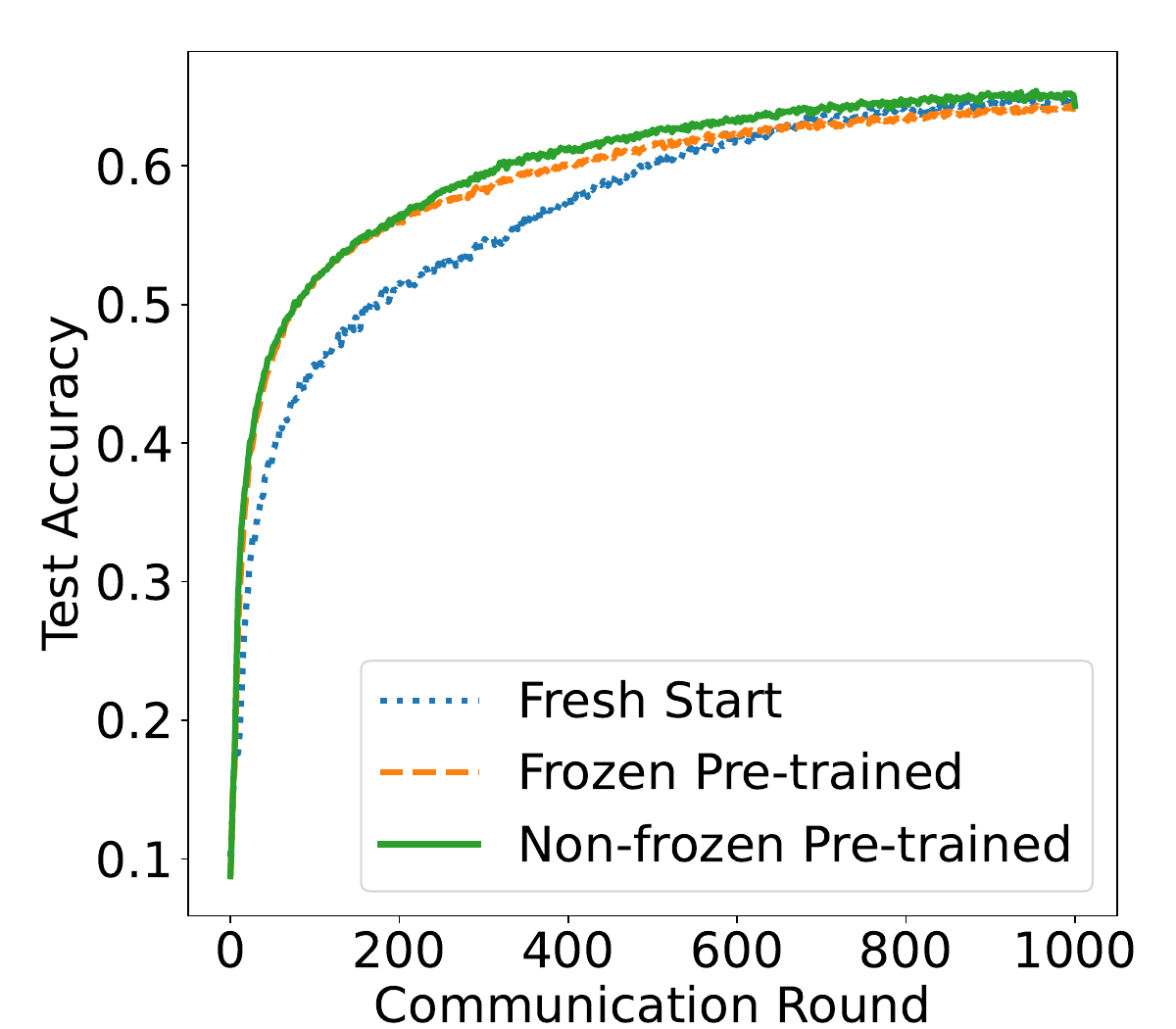}%
    \caption{The training curves utilizing the pre-trained embedding model. ``Fresh Start'' denotes that the model is trained from scratch without using the pre-trained model. ``Frozen Pre-trained'' signifies that the pre-trained model is used but remains unchanged during training. ``Non-frozen Pre-trained'' indicates that the pre-trained model is used and updated during training.}
    \label{fig:pretrain}
\end{figure}

To further reduce the amount of server-client communication, we leverage pre-trained embedding models in the experiments.
As shown in Fig.~\ref{fig:pretrain}, utilizing pre-trained embedding models accelerates convergence significantly. Additionally, the use of frozen pre-trained embedding models, although slightly degrading the model's performance, eliminates the need for embedding model aggregation, thus effectively reducing the amount of communication between the server and clients.

\section{conclusion}
\label{sec:conclusion}
This paper introduces \namenospace, a distributed MoE training algorithm that leverages the relationship between gating network parameters and expert selection patterns to capture correlations among experts across clients, enhancing model robustness and personalizability.
Additionally, \name alleviates server-client communication pressure through P2P expert model transmission. 
Experimental results demonstrate that \name significantly improves performance and reduces server-client transmission.
Furthermore, employing pre-trained embedding models further decreases server-client transmission without compromising performance.
We believe that \name can support varying numbers of experts and accommodate heterogeneous expert models through knowledge distillation, better addressing device heterogeneity in FL systems.

\bibliographystyle{IEEEtran}
\bibliography{main}

\end{document}